\documentclass{article}

\usepackage{PRIMEarxiv}
\usepackage{amsmath}
\usepackage[utf8]{inputenc} % allow utf-8 input
\usepackage[T1]{fontenc}    % use 8-bit T1 fonts
\usepackage{hyperref}       % hyperlinks
\usepackage{url}            % simple URL typesetting
\usepackage{booktabs}       % professional-quality tables
\usepackage{amsfonts}       % blackboard math symbols
\usepackage{nicefrac}       % compact symbols for 1/2, etc.
\usepackage{microtype}      % microtypography
\usepackage{lipsum}
\usepackage{fancyhdr}       % header
\usepackage{graphicx}       % graphics
\graphicspath{{media/}}     % organize your images and other figures under media/ folder
\usepackage{array}  

\title{Automatic Simplification of Common Vulnerabilities and Exposures Descriptions
}

\author{
  Varpu Vehomäki, Kimmo Kaski \\
  Department of Computer Science \\
  Aalto University School of Science \\
  Espoo\\
  Finland \\
  \texttt{firstname.lastname@aalto.fi}}

\begin{document}
\maketitle

\begin{abstract}
Understanding cyber security is increasingly important for individuals and organizations. However, a lot of information related to cyber security can be difficult to understand to those not familiar with the topic. In this study, we focus on investigating how large language models (LLMs) could be utilized in automatic text simplification (ATS) of Common Vulnerability and Exposure (CVE) descriptions. Automatic text simplification has been studied in several contexts, such as medical, scientific, and news texts, but it has not yet been studied to simplify texts in the rapidly changing and complex domain of cyber security. We created a baseline for cyber security ATS and a test dataset of 40 CVE descriptions, evaluated by two groups of cyber security experts in two survey rounds. We have found that while out-of-the box LLMs can make the text appear simpler, they struggle with meaning preservation. Code and data are available at https://version.aalto.fi/gitlab/vehomav1/simplification\_nmi.
\end{abstract}

% keywords can be removed

\section{Introduction}
Text simplification is the process of re-writing a natural language text in a less complicated manner, while preserving the original meaning of the text. For the rapidly increasing and specialised text corpora of today, there is growing need to perform this process automatically or semi-automatically using Artificial Intelligence (AI) and Large Language Models (LLMs). This process is called Automatic Text Simplification (ATS). 

Text simplification is often used to make information accessible to language learners, people with cognitive disabilities, or those without sufficient contextual expertise, which divides the goal of text simplification into two main styles, namely Plain Language and easy-read \cite{freyer_easy-read_2024}. Plain Language is aimed at people without domain expertise, while easy-read text also attempts to be understandable for those with cognitive disabilities \cite{freyer_easy-read_2024}. Here we will focus on the Plain Language approach while leaving the easy-read approach outside the scope of the present investigation.

Some of the fields of research in which ATS is actively studied are scientific \cite{ali_team_nodate, ScientificTextSimplificationMinimumBayes} and medical  \cite{van_den_bercken_evaluating_2019, ondov2022survey} texts, but not yet in the complex context of cyber security. It often happens that even experts in one field of research will have difficulty understanding research articles somewhat outside of their own field \cite{ermakova2023overview}, which has recently served as motivation for many research-related ATS studies. To make interdisciplinary communication easier, the authors of \cite{guo_personalized_2024} studied making texts more understandable through the detection of personalized jargon. Their system takes into account the background of the user and modifies the text to make it more easily understandable. It was found that those less familiar with the domain preferred term explanations and that those with some more knowledge found contextualizing complex terminology useful.

As it comes to using LLMs in ATS, it should be noted that they can also cause harm even when used for non-malicious purposes. LLMs' biases can also show in text simplification and, for certain groups of end users, simplified text may turn out to be misleading \cite{freyer_easy-read_2024}. If important information is incorrectly simplified, there is a risk of harm, which without proper human oversight can result in a responsibility gap and thus needing extra attention to be paid in the ATS process. For example, a question can be raised that who is held responsible if an ATS system omits a key detail in a simplified version of a cyber security report? In addition to the audience of the simplified reports possibly being misled, such incidents can erode people's trust in the ATS system and in the organization using it.

To avoid these problems in text simplification, collaboration with the intended audience, domain experts, and developers has been proposed \cite{freyer_easy-read_2024}. One of the problems is factuality in commonly used text simplification corpora and in the output of text simplification models. It has also been shown that the metrics used to evaluate the performance of text simplification models do not capture factuality \cite{freyer_easy-read_2024}. In addition to human oversight, collaborative systems and transparency about the use of ATS are ways to make information more reliable \cite{devaraj2022evaluating}.

Responsibility in text simplification does not only cover the possible mistakes in the simplified text. Certain ATS solutions can shift the responsibility of understanding to the reader. It is important that there is an attempt to write accessible source text and that ATS is only used as a tool to assist in writing text so that it is accessible to a wider audience. Writing original texts in complex language because an ATS system is available for the intended audience to use should not be the goal \cite{kim_considering_2024}.

In todays cyber security attacks, written reports of incidents are an important part of a security analyst's job. These reports bring together important information on attempted and successful attacks against businesses, infrastructures, and other entities. However, these reports tend to be technical and thus hard to understand for those who are not experts in cyber security. This poses a major challenge especially to strategy level decision makers of e.g. infrastructure organisations. However, making a simplified version of such a report can be time-consuming and challenging for someone not familiar with the task. In addition, a rapid increase in the number of reports of cybersecurity incidents, vulnerabilities, and exploits calls for the development of methods or tools for automatic text simplification (ATS). 

In order to make these cyber security reports better accessible for non-expert audiences, we introduce an AI-assisted ATS method for simplifying descriptions of Common Vulnerabilities and Exposures (CVE) descriptions. Furthermore, we create a small human-evaluated semi-synthetic dataset of simplified CVE descriptions to be used for testing and comparison purposes of ATS systems.

\section{Materials and methods}
For the dataset to be used in this study, we randomly chose 100 Common Vulnerability and Exposure (CVE) descriptions published in 2025 from the CVElistV5 repository\footnote{https://github.com/CVEProject/cvelistV5}. Of these 100 CVE descriptions, we chose 40 for human evaluation, and the remaining 60 descriptions were used as an evaluation set in the development of the local simplification model. Before simplifying the 40 CVEs in the human evaluation set, we manually cleaned up the CVE descriptions, i.e., removing log excerpts and other non-natural language texts. We excluded non-natural language because it is outside the scope of this study.

The test set of 40 CVEs was first simplified with the GPT4o model, using the Azure OpenAI API\footnote{https://learn.microsoft.com/en-us/azure/ai-foundry/openai/overview}. These initial simplifications were made in April 2025 using the GPT version gpt-4o-2024-11-20. After the first round of human evaluation, we resimplified these 40 CVE for the second round of human evaluation. These resimplifications were performed using the ChatGPT GUI to enable real-time interaction with the model. Here we used the same model as in the case of initial simplifications, namely GPT-4o, and these resimplifications were performed in July 2025.

The initial simplifications were performed at the sentence level, as many of the evaluation metrics commonly used in ATS are based on it. The CVEs were split into sentences using the sentencizer from the spaCY library\footnote{https://spacy.io/api/sentencizer}. Then these sentences were given to the GPT4o model one by one, and then the simplification was aligned with the original sentence. Subsequent improvements to the initial simplifications were made at the document level. The reason for this was that the human evaluation was done at the document level, as assessing the quality and meaning preservation is more easily done at the document level than at the sentence level.

As commercial LLMs lack transparency and user control, we tested and trained some open-source models specifically for this task. We experimented with several different approaches, like training a BART model using domain-adaptive pre-training (DAPT) and running open-source LLMs locally. After some initial testing, we settled on an agent system using the 4 billion parameter version of Gemma 3 \cite{team_gemma_2025}. We chose this model mainly for its good performance in relation to size, as it can be run on a single GPU system yielding a coherent output.

\begin{figure}[h]
  \centering
  \includegraphics[width=5cm]{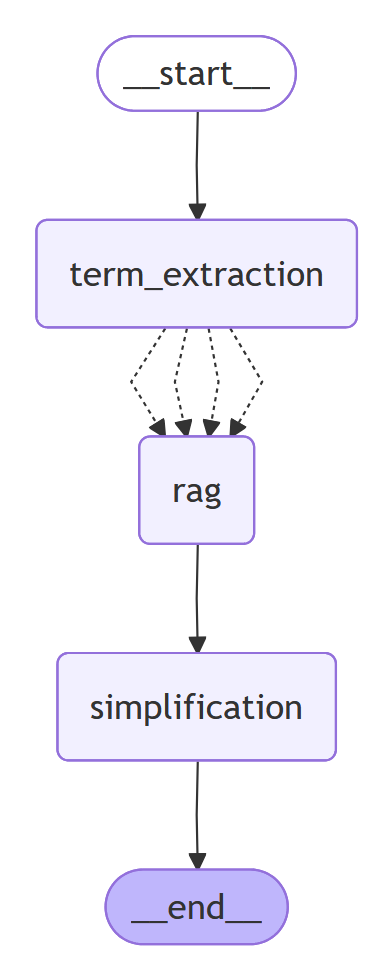}
  \caption{A figure illustrating the structure of the GemmaAgent.\label{agent_graph}}
\end{figure}

Figure \ref{agent_graph} shows the pipeline structure of the agent-based Gemma model system or, in short, GemmaAgent. As a first step of the process, the original CVE is fed into a term extraction agent that outputs a list of terms in the text. For the term extraction stage, we experimented with the Gemma base model and several named entity recognition (NER) models. For the final system to extract terms, we used the AITSecNER\footnote{https://huggingface.co/selfconstruct3d/AITSecNER} model. Here we restricted the term extraction to terms with labels 'CON', 'MALWARE', 'TACTIC', 'TECHNIQUE' or 'TOOL', because we found these categories to be most relevant for the understandability of the CVE. The terms extracted by the NER model are then fed into a retrieval augmented generation (RAG) agent with access to a database containing cybersecurity lexica and dictionaries. The RAG agent writes explanations of the extracted terms based on the information found in the database. Finally, the simplification agent receives the original CVE description along with the term explanations and instructions to simplify the original CVE using the term explanations as support.

\subsection{Evaluation}

Several metrics have been created to assess the performance of text simplification methods. One of them is SARI
\cite{xu_optimizing_2016} which is a sentence-level metric to evaluate
the quality of a simplified sentence by comparing it to the original sentence and a simplified reference sentence. In this paper, we use D-SARI, a version of SARI to evaluate document-level text simplification \cite{sun_document-level_2021}. Like the original SARI metric, D-SARI requires a reference to perform the evaluation. The formula for D-SARI is the following:

\[
\text{D-SARI} = \frac{D_{\text{keep}} + D_{\text{del}} + D_{\text{add}}}{3}
\]
where $D_{keep}$, $D_{add}$ and $D_{del}$ are the addition, deletion and keep operations defined in \cite{xu_optimizing_2016} multiplied by the penalty scores defined in \cite{sun_document-level_2021}.

The Flesch-Kincaid Grade Level (FKGL) is another common ATS evaluation metric \cite{flesch1948new}, which is calculated as follows 

 \begin{align*}
 \text{FKGL} = 0.39 \cdot \text{ASL} + 11.8 \cdot \text{ASW} - 15.59.    
 \end{align*}
where ASL = average sentence length in words and ASW = average number of syllables per word.
In \cite{tanprasert_flesch-kincaid_2021} it was argued that FKGL is not a suitable metric for automatic text simplification and that it is better to separately report the average sentence length and the average number of syllables, as they are the components of FKGL.

The other important dimension of text simplification is the preservation of meaning, which can be evaluated, for example, by using BERTScore \cite{zhang_bertscore_2020} and semantic similarity. 
The BERTScore is calculated as follows:

\[
P = \frac{1}{m} \sum_{i=1}^{m} \max_{j \in [1,n]}
\cos \left( \mathbf{h}(x_i), \mathbf{h}(y_j) \right)
\]

\[
R = \frac{1}{n} \sum_{j=1}^{n} \max_{i \in [1,m]}
\cos \left( \mathbf{h}(x_i), \mathbf{h}(y_j) \right)
\]

\[
\text{BERTScore}_{F1} = \frac{2PR}{P+R}
\]
where P = token level precision, R = token level recall and h = embedding function. For more details on the BERTScore formula, refer to \cite{zhang_bertscore_2020}.
Semantic similarity is yet another metrics to measure how close two pieces of text are in meaning. The formula for the semantic similarity is as follows
\[
\text{Semantic similarity}(s_c, s_r) =
\cos \left( \mathbf{e}(s_c), \mathbf{e}(s_r) \right)
\]
where e = embedding function, $s_c$ = candidate sentence, $s_r$ = reference sentence and $cos(x,y)$ = cosine similarity between vectors x and y. We use two different embedding models and libraries to calculate semantic similarity: MeaningBERT \cite{beauchemin_meaningbert_2023} and Sentence-BERT \cite{reimers_sentence-bert_2019} with the nomic-embed-text-v1.5\footnote{https://huggingface.co/nomic-ai/nomic-embed-text-v1.5} model.

In this study we have also performed evaluations of the 40 randomly chosen CVE descriptions, simplified by GPT-4o, in the form of a human survey using a 3-point Likert scale. We conducted two rounds of human surveys: the first round with 12 participants and the second round with 10 participants, all with some level of cyber security expertise.

Participants in the first survey were shown the original and simplified CVE descriptions and asked to respond to two statements: "\textit{The simplification is easier to understand than the original}" and "\textit{The simplification preserves the meaning of the original text}". The respondents were given the same three response options for both statements: agree, neither agree nor disagree, and disagree. There was also an optional comment field where respondents could add comments about the simplifications.

The first round or initial simplifications that received more than 80\% "agree" responses and no "disagree" responses to both statements above were automatically considered high-quality and were left out of the second human evaluation round. For the second round of simplifications, we prompted the GPT-4o model to improve the initial simplifications. We presented the model with the original prompt, instructions on what to do next, the original CVE, the initial simplification, and the comments given on each CVE by the participants of the first survey.

In the second round of human survey, we asked the participants to evaluate the quality of the first and second text simplification versions and also compare the first and second round simplification, adding the statement: \textit{"The second round simplification is of better quality than the first"}.

\section{Results}

Table \ref{tab:sari} shows the results of the D-SARI scores for the GPT4o, Gemma3:4b, and GemmaAgent models. For all of these models, the scores turned out to be overall low, but among them Gemma3:4b got the highest score of 0.14. As D-SARI is the only reference-based metric used, the low scores can be attributed to differences between the simplifications and the references. However, we expect a much larger number of verified reference simplifications to improve the values of these scores.

Table \ref{tab:meaning} shows the results of BERTScore, MeaningBERT, and SentenceBERT scores for the Semi-synthetic, GPT4o, Gemma3:4b, and GemmaAgent models. It turned out that Semi-synthetic and GemmaAgent simplifications performed best on the meaning preservation metrics. The semi-synthetic data got the best BERTScore, while GemmaAgent got a slightly higher MeaningBERT score. The semantic similarity values for both models were the same. Gemma 3:4b got the lowest BERTScore and Semantic similarity score, so we can conclude that adding retrieval augmented generation (RAG) in the simplification process helps with meaning preservation. The MeaningBERT score for Gemma3:4b was slightly higher than for GPT-4o.

Table \ref{tab:simplicity} shows the results for the system level FKGL scores as well as the average word, syllable, and sentence counts of the original and simplified CVEs. For the calculations of these scores, we used the lftk-library\footnote{https://github.com/brucewlee/lftk}. The simplifications produced by GPT-4o achieved somewhat better scores for referenceless automatic simplicity metrics than the original descriptions. The FKGL score for the original descriptions was 12.45, while GPT-4o reduced the score to 9.49. Gemma3:4b shortened the descriptions, based on the average number of words and sentences, while the other models produced simplifications that were longer than the original text. In addition, we calculated the average number of named entities for the original and simplification CVEs, using the AITSecNER NER model. All other models turned out to reduce the number of named entities, except the GemmaAgent model.

\begin{table}[htp]
    \centering
    \begin{tabular}[width=0.7\columnwidth]{| c | c |}
    \hline
    Model  & D-SARI \\
    \hline
    GPT-4o & 0.09  \\
    Gemma3:4b & \textbf{0.14} \\
    GemmaAgent & 0.09\\
    \hline
    \end{tabular}
    \caption{System level D-SARI scores, test set results.}
    \label{tab:sari}
\end{table}

\begin{table}[htp]
\small
    \centering
    \begin{tabular}[width=0.7\columnwidth]{| c | c | c | c |}
    \hline
    Model  & BERTScore & MeaningBERT & SBERT\\
    \hline
    Semi-synthetic & \textbf{0.60} & 76.62 & \textbf{0.92} \\
    GPT-4o & 0.58 & 71.66 & 0.90\\
    Gemma3:4b & 0.50 & 72.80 & 0.86\\
    GemmaAgent & 0.56 & \textbf{76.94} & \textbf{0.92}\\
    \hline
    \end{tabular}
    \caption{System level scores for meaning preservation, test set results.}
    \label{tab:meaning}
\end{table}

\begin{table*}[htp]
    \centering
    \begin{tabular}[width=0.7\columnwidth]{| m{2.25cm} | m{1cm} | m{2.5cm} | m{2.5cm} | m{3cm}| m{3cm}|}
    \hline
    Model  & FKGL & Average number of words & Average number of sentences & Average number of syllables per word & Average number of named entities\\
    \hline
    Original & 12.45 & 43.68 & 2.08 & 1.62 & 2.6\\
    Semi-synthetic & 10.74 & 59.58 & 2.55 & 1.39 & 2.23\\
    GPT-4o & \textbf{9.49} & 54.85 & 2.53 & 1.35 & 1.65\\
    Gemma3:4b & 14.50 & 34.65 & 1.3 & 1.61 & 1.33 \\
    GemmaAgent & 14.20 & 55.3 & 2.1 & 1.61 & 2.85\\
    \hline
    \end{tabular}
    \caption{System level scores for simplicity, test set results.}
    \label{tab:simplicity}
\end{table*}

In addition, it turned out that out of 40 CVE simplifications, only five simplifications were accepted from the first round of human evaluations without modifications to the second round. Based on these evaluations, the remaining 35 were modified and passed on to the second round of human evaluation.

\section{Discussion}

In the first round of simplifications, it was found out that the GPT-4o would not return simplifications for certain types of prompt, like "Exploitation of this issue requires user interaction in that a victim must open a malicious file". We suspect that this is caused by GPT-4o guardrails that prevent it from outputting the so-called "harmful content". To ensure that the evaluations would not be affected by such a technical issue, the prompts with which a simplification was not returned were manually substituted with the original sentence before the evaluations. Furthermore, GPT-4o turned out to be the only model in our experiments that reduced the FKGL scores of the descriptions from the originals', reaching even a better FKGL score than the semi-synthetic data. The GPT-4o produced longer simplifications with more sentences and less syllables per word than the other fully automatic systems, which explains the lower FKGL score.

Although GPT-4o mostly preserved information that should be left untouched, such as version numbers, without specific instructions to do so, there was at least one instance in which the model did not correctly deal with the version number. For example, in CVE-2025-32202\footnote{https://nvd.nist.gov/vuln/detail/CVE-2025-32202} GPT-4o incorrectly rounded the version number from 4.3000000025 to 4.3. Stating that version information should be left untouched in the prompt could help avoid problems like this.

One of the interesting findings is that the GemmaAgent model performed well in the meaning preservation metrics, but did not lower the complexity of the text, based on the metrics in Table \ref{tab:simplicity}. In addition, it turned out to be the only model producing a higher average number of named entities than the original CVEs. This could be caused by the focus on explaining terms in the prompt given to the model.

The semi-synthetic data went through multiple iterations, and the simplifications were improved on the basis of the feedback from domain experts, as can be expected. These referenceless automatic metrics show that the test set we have produced turned out to be a good reference for ATS testing in cybersecurity. The biggest challenge in the simplifications turned out to be the preservation of meaning, which caused the most disagreement between human evaluators.

One of the key observations of this study is that it is important to consider the audience or end users to whom the simplifications are aimed. Here, our objective was to produce text that would be generally understandable to people with limited understanding of the cyber security domain. We found that even within an audience, there is variation on the level of understanding and that needs to be taken into consideration when developing text simplification systems. Equally important is that, if the simplifications produced by LLMs are used in the decision-making process of a company, it is crucial that the information in the simplified report is factual and that no important points are left out.

\section{Conclusions}

Our study shows that automatic evaluation metrics do not consistently follow human understanding of text simplicity. It has also shown that meaning preservation is a big challenge in ATS. Some of the problems with meaning preservation can be dealt with by improving prompting, but there are also aspects of meaning that are not easily measured and thus are harder to evaluate.

More research is needed on human-in-the-loop simplification systems and proper human evaluations of these systems. Our surveys were targeted towards cyber security professionals, but it can be very challenging for them to evaluate how understandable the text would be to someone without their background, while a non-expert will not be able to reliably evaluate the meaning preservation.

A larger database of cyber security information could improve the performance of the RAG model and make simplifications more reliable. More detailed investigation of the effects of RAG in ATS is needed.

\subsection{Conflict of interest}
The authors of this study declare no conflict of interest.

\bibliographystyle{unsrt}  
\bibliography{references}

\end{document}